\documentclass[conference]{IEEEtran}
\IEEEoverridecommandlockouts
\usepackage{cite}
\usepackage{amsmath,amssymb,amsfonts}

\usepackage{graphicx}
\usepackage{textcomp}
\usepackage{xcolor}

\usepackage{algorithm}
\usepackage{algorithmicx}
\usepackage{algpseudocode}

\usepackage{eqparbox}
\newdimen{\algindent}
\algnewcommand\LeftComment[2]{\hspace{#1\algindent} \eqparbox{COMMENT}{#2} \hfill}
\usepackage{multirow}
\usepackage{booktabs}
\usepackage{cleveref}
\Crefname{section}{Section}{Sections}
\crefname{section}{Sec.}{Secs.}
\Crefname{table}{Table}{Tables}
\crefname{table}{Tab.}{Tabs.}
\Crefname{figure}{Figure}{Figures}
\crefname{figure}{Fig.}{Figs.}
\Crefname{equation}{Equation}{Equations}
\crefname{equation}{Eq.}{Eqs.}

\def\BibTeX{{\rm B\kern-.05em{\sc i\kern-.025em b}\kern-.08em
    T\kern-.1667em\lower.7ex\hbox{E}\kern-.125emX}}
\begin{document}

\title{DynamicTrack: Advancing Gigapixel Tracking in Crowded Scenes
\thanks{$^*$Co-first author, $^\dag$Corresponding author.}
\thanks{This work is supported in part by National Natural Science Foundation of China (NSFC) under contract No. 62125106, 61860206003, 62088102 and U21B2013.}
}

\author{
\IEEEauthorblockN{Yunqi Zhao$^*$}
\IEEEauthorblockA{
\textit{Tsinghua University}\\
Beijing, China \\
yq-zhao22@mails.tsinghua.edu.cn}
\\
\IEEEauthorblockN{Kai Ni}
\IEEEauthorblockA{
\textit{HoloMatic Technology Co.,Ltd.}\\
Beijing, China \\
nikai@holomatic.com}
\and
\IEEEauthorblockN{Yuchen Guo$^*$}
\IEEEauthorblockA{
\textit{Tsinghua University}\\
Beijing, China \\
yuchen.w.guo@gmail.com}
\\
\IEEEauthorblockN{Ruqi Huang}
\IEEEauthorblockA{
\textit{Tsinghua University}\\
Beijing, China \\
ruqihuang@sz.tsinghua.edu.cn}
\and
\IEEEauthorblockN{Zheng Cao}
\IEEEauthorblockA{
\textit{Biren Technology}\\
Beijing, China \\
zcao@birentech.com}
\\
\IEEEauthorblockN{Lu Fang$^\dag$}
\IEEEauthorblockA{
\textit{Tsinghua University}\\
Beijing, China \\
fanglu@tsinghua.edu.cn}
}

\maketitle

\begin{abstract}
Tracking in gigapixel scenarios holds numerous potential applications in video surveillance and pedestrian analysis. Existing algorithms attempt to perform tracking in crowded scenes by utilizing multiple cameras or group relationships. However, their performance significantly degrades when confronted with complex interaction and occlusion inherent in gigapixel images.
In this paper, we introduce DynamicTrack, a dynamic tracking framework designed to address gigapixel tracking challenges in crowded scenes.
In particular, we propose a dynamic detector that utilizes contrastive learning to jointly detect the head and body of pedestrians. Building upon this, we design a dynamic association algorithm that effectively utilizes head and body information for matching purposes. 
Extensive experiments show that our tracker achieves state-of-the-art performance on widely used tracking benchmarks specifically designed for gigapixel crowded scenes.
\end{abstract}


\section{Introduction}
\par As the development of imaging devices, the acquisition of gigapixel images\cite{Brady2012Multiscale, Yuan2021Modular} has become increasingly feasible. Gigapixel images, with large spatial coverage and high imaging quality, have substantial potential applications in smart cities, such as traffic monitoring\cite{Tang2019Cityflow} and pedestrian surveillance\cite{Cheng2011HumanGA} in crowded scenes.
\par Although gigapixel images offer richer semantic information and finer-grained targets for crowded scenes compared to megapixel images, they also introduce complex interactions and severe occlusion issues, posing new challenges for tracking. 
Some researchers attempt to address severe occlusion in crowded scenes through multi-camera tracking\cite{Eshel2010Tracking}. However, the rigid separation of continuous space by multiple cameras leads to the dispersion of spatial information. 
Others seek robust tracking by leveraging interaction information provided by group relationships\cite{Zhu2014Crowd} among pedestrians, yet capturing group relationships in crowded scenes proves challenging. 
The existing issues with current methods make it difficult to apply gigapixel images in practice.

\begin{figure}[t]
        \centering
	\includegraphics[width=0.45\textwidth]{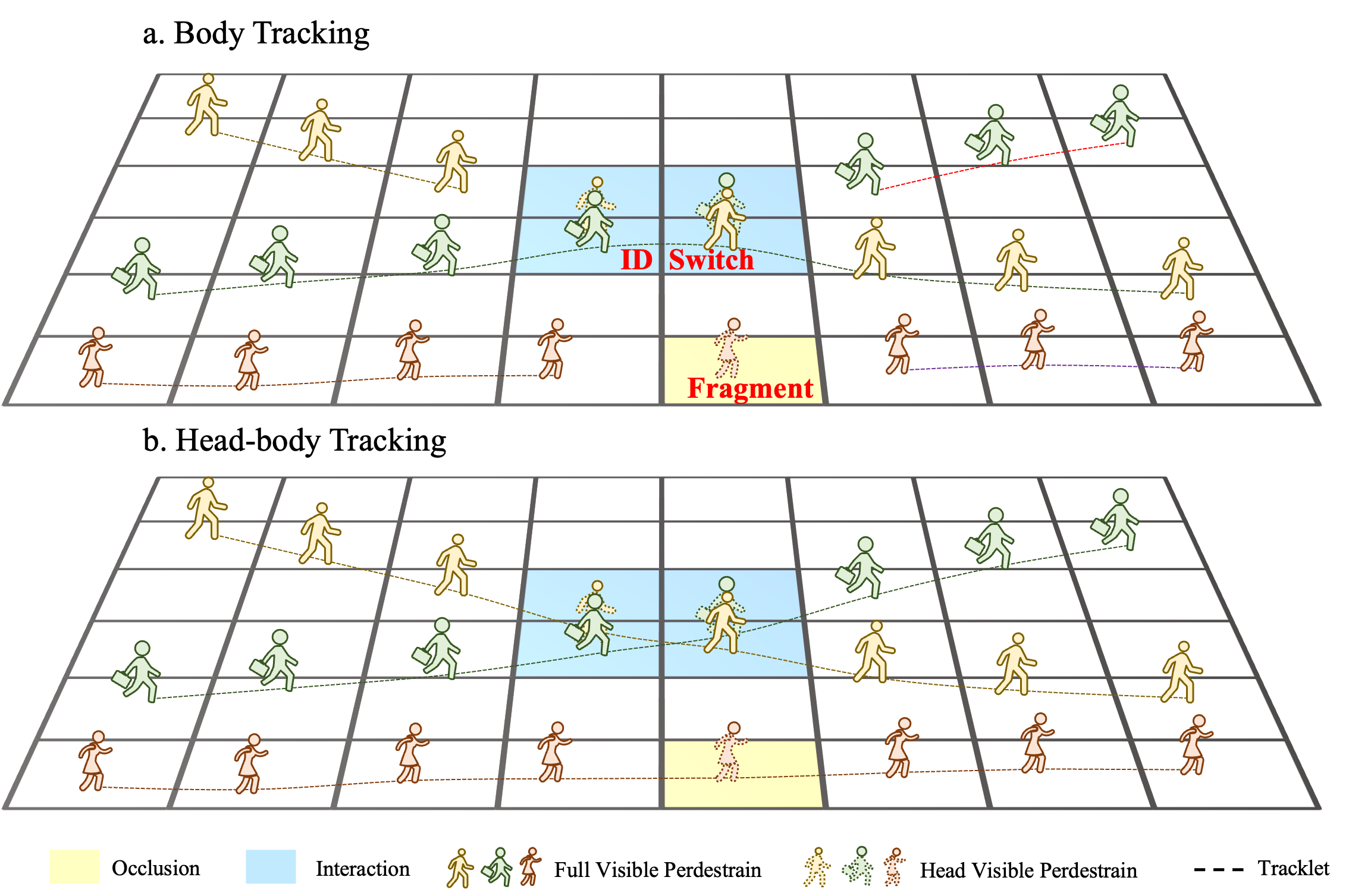}
	\caption{The comparison between body tracking and head-body tracking: a. Body tracking encounters ID switch and fragment in interactive and occluded scenarios. b. Head-body tracking is robust in crowded scenes.}
	\label{fig:teaser}
        \vspace{-0.3cm}
\end{figure}

\par In this paper, we present DynamicTrack that comprises dynamic detection and association modules to address the above mentioned challenges in gigapixel crowded scenes. 
Head features are less likely to be obscured and thus provide robust trajectory cues, enhancing tracking accuracy in crowded scenes. Based on this observation, we propose a dynamic detector for head-body joint detection, facilitating the joint tracking of heads and bodies. Specifically, we incorporate embedding learning into the dynamic detector and utilize the embedding loss derived from contrastive learning for feature learning. 
To fully leverage the distinct characteristics of the head and body in crowded scenarios, we propose a dynamic association algorithm that incorporates head features into the matching process. The dynamic association algorithm treats the body as the core and the head as the support, which combines fine-grained local head features with global body information. 
We conduct extensive experiments to evaluate the performance of our proposed tracker. The results demonstrate that our dynamic detector achieves the best performance in joint detection on the Crowdhuman\cite{shao2018crowdhuman} dataset. Moreover, our proposed DynamicTrack outperforms state-of-the-art methods on MOT20\cite{Dendorfer2020MOTChallengeAB} and PANDA\cite{wang2020panda} datasets.
\par Our contributions are summarized as the following:
\begin{enumerate}
    \item Dynamic Detection: 
    We propose a dynamic detector based on contrastive learning, which enables joint head and body detection for gigapixel tracking.
    \item Dynamic Association: 
    We introduce a dynamic association algorithm to fully exploit the potential of both head and body cues for joint matching.
    \item We demonstrate the superior performance of our tracker on widely used tracking benchmarks designed for crowded scenes.
\end{enumerate}

\section{Related Work}
With the development of imaging devices, obtaining gigapixel images\cite{Brady2012Multiscale, Yuan2021Modular} is no longer difficult. Wang et al. introduced the PANDA\cite{wang2020panda} dataset, designed to address visual tasks under gigapixel resolution. Although PANDA offers rich semantic annotations and fine-grained information, its crowded scenes introduce complex interaction and occlusion, posing significant challenges to tracking. In this work, we incorporate joint head-body tracking into the tracking framework, which enables robust gigapixel tracking in crowded scenes.
Some studies have attempted joint detection of head and body. PedHunter~\cite{chi2020pedhunter} employs a mask-guided module to leverage the head information to enhance the representation learning of pedestrian features. Double Anchor R-CNN~\cite{zhang2019doubleanchor} presents a double-anchor RPN to capture body and head parts in pairs. JointDet~\cite{chi2020jointdet} detects head and body simultaneously and performs relational learning between them for joint detection. These methods indirectly utilize head information for better detection but have limitations in achieving end-to-end optimization.
Some researchers have attempted to integrate head information into tracking frameworks. Sun et al.\cite{sun2021improved} utilize the harder-to-obscure head feature as the basis for tracking and replace the association result with the matching body. However, tracking based on the head is not suitable for tasks focused on the body as the target. Zhang et al.\cite{zhang2023handling} perform head-body matching based on a positional prior followed by joint head-body tracking. However, the relative position prior of the head and body is not robust under occlusion.

\begin{figure*}[t]
	\includegraphics[width=\textwidth]{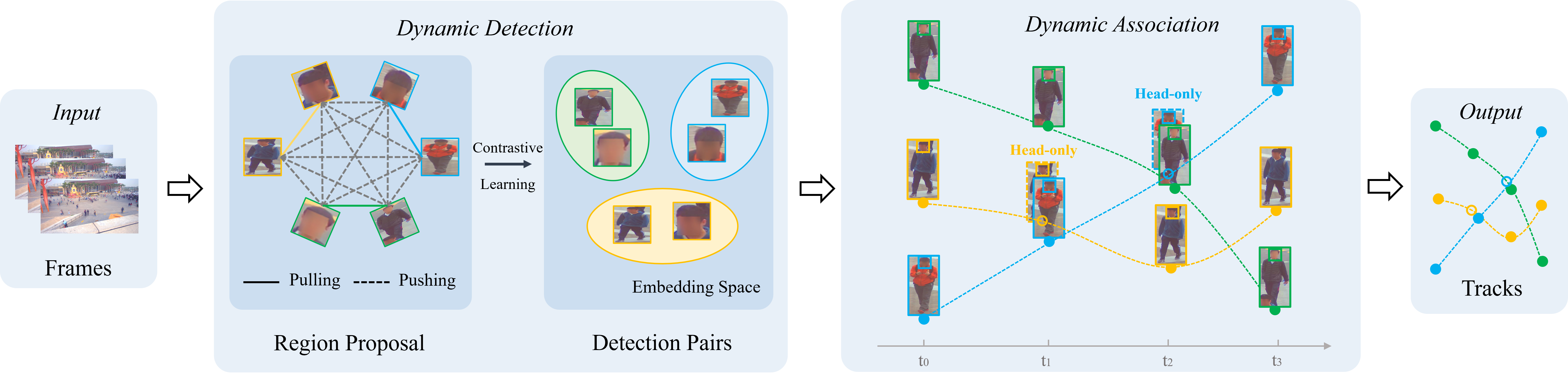}
	\caption{Overview of DynamicTrack framework for gigapixel tracking.
		\textit{Dynamic Detection:} 
            Contrastive learning-based detector achieves simultaneous detection of both the body and the head for pedestrian tracking.
		\textit{Dynamic Association:} 
            Dynamically utilizing head and body of the same identity for matching to achieve robust tracking in crowded scenes."} 
	\label{fig:pipeline}
        \vspace{-0.2cm}
\end{figure*}
\begin{figure}[t]
        \centering
	\includegraphics[width=0.45\textwidth]{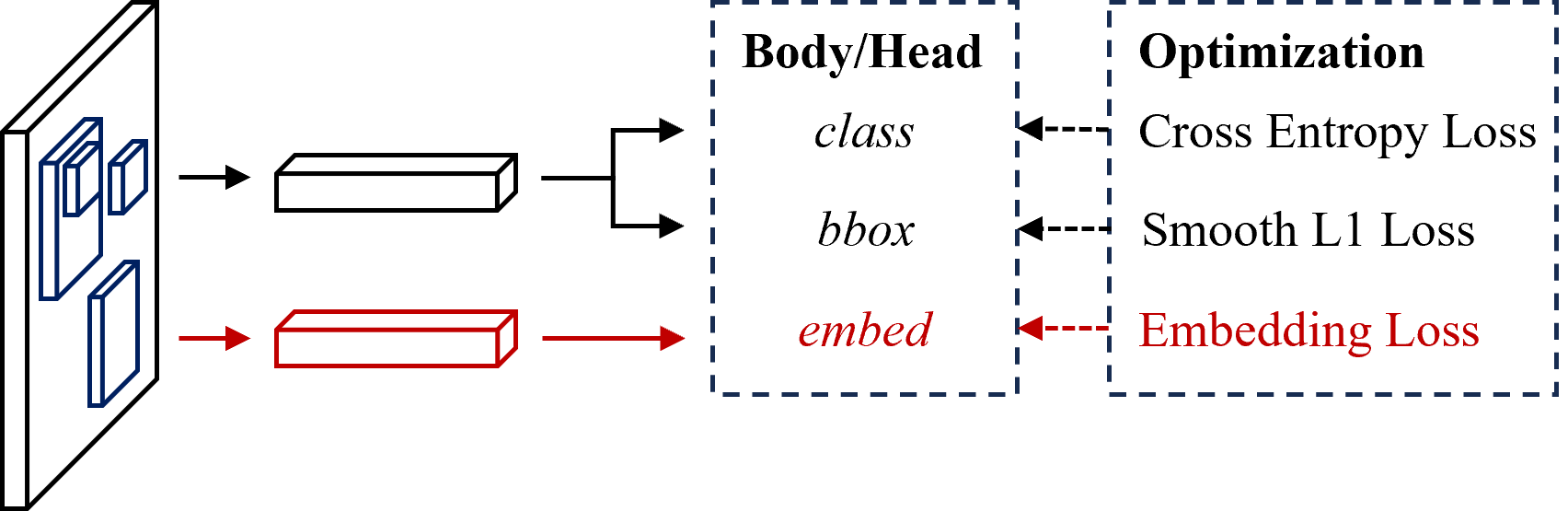}
	\caption{The framework of our dynamic detector for head-body detection consists of a modified version of the classical two-stage detector, Faster-RCNN\cite{ren2015faster}. We introduce an additional branch for embedding learning and leverage an associative embedding loss based on contrastive learning for supervision.}
	\label{fig:detector}
        \vspace{-0.2cm}
\end{figure}

\section{DynamicTrack}
\par Our goal is to design a gigapixel tracker for crowded scenes. We introduce a dynamic tracking framework into the traditional Separate Detection and Embedding tracking framework\cite{wojke2017deepsort, zhang2022bytetrack} and the proposed \textit{DynamicTrack} framework is shown in ~\cref{fig:pipeline}. 
First, we implement an end-to-end dynamic detector based on contrastive learning which is capable of detecting both the body and head of a pedestrian. Then, we propose a dynamic association algorithm that can simultaneously utilize head and body features for robust tracking. 

\subsection{Dynamic Detection}
\label{sec:dynamic_detection}
\par The key challenge in gigapixel tracking is dealing with crowded scenes that involve complex interaction and occlusion among pedestrians. To tackle this issue, we have developed an end-to-end dynamic detector capable of simultaneously capturing both the head and body of a pedestrian. Our approach is based on the understanding that head features are less prone to occlusion in crowded environments, and they can thus offer more comprehensive and reliable features for subsequent tasks.
To achieve joint detection, we draw inspiration from the concept of associative embedding learning~\cite{hadsell2006dimensionality} and utilize the associative embedding technique to establish the relationship between the head and body of a pedestrian. As illustrated in ~\cref{fig:detector}, we incorporate a parallel branch into the Faster R-CNN~\cite{ren2015faster} framework. This additional branch is placed after the ROI feature and functions as the embedding module, producing an embedding vector for each instance.  To optimize this embedding module, we introduce an Associative Embedding Loss (AML). The AML aims to encourage embeddings from the same pedestrian to be pulled closer together while pushing apart the embeddings belonging to different individuals.
\par \noindent \textbf{Preliminaries:} Given the ground truth annotations $\mathcal{G}$: 
\begin{equation}
   \mathcal{G}=
   \left\{
   (g_{n}^{(b)}, g_{n}^{(h)})
   \mid
   g_{n}^{(b)} \in \mathcal{G}^{(b)},
   g_{n}^{(h)} \in \mathcal{G}^{(h)}
   \right\}
\end{equation}
where $\mathcal{G}^{(b)}$ represents the set of body boxes, $\mathcal{G}^{(h)}$ represents the set of head boxes, and $(g_{n}^{(b)}, g_{n}^{(h)})$ represents the matched body and head pair. 
And the predicted body set is $(\mathcal{D}^{(b)}, \mathbf{e}^{(b)})$ and head set is $(\mathcal{D}^{(h)}, \mathbf{e}^{(h)})$, where $\mathcal{D}$ represents the detection results and $\mathbf{e}$ represents the corresponding embedding features.

\par \noindent \textbf{Pulling Loss:} 
To ensure that the embedding vectors of positive pairs are close to each other, we design pulling losses for various cases: body and body ($bb$), head and head ($hh$), and matched body and head ($bh$). The pulling loss functions for three cases are as follow:
\begin{equation}
\left\{
\begin{array}{l}
L_{pull}^{bb}=\frac{1}{M^{2}}\sum\limits_{i=1}^{M}\sum\limits_{j=1, j \neq i}^{M} e^{d_{i j}} \| \mathbf{e}_{i}^{(b)} \mathbf{e}_{j}^{(b)} \|^{2} \\
L_{pull}^{hh}=\frac{1}{N^{2}}\sum\limits_{i=1}^{N}\sum\limits_{j=1, j \neq i}^{N} e^{d_{i j}} \| \mathbf{e}_{i}^{(h)} \mathbf{e}_{j}^{(h)} \|^{2}  \\
L_{pull}^{bh}=\frac{1}{M} \frac{1}{N}\sum\limits_{i=1}^{M}\sum\limits_{j=1}^{N} \| \mathbf{e}_{i}^{(b)} \mathbf{e}_{j}^{(h)} \|^{2}  
\end{array}
\right.
\end{equation}
where $M$ and $N$ represent the number of the predicted body set $\mathcal{D}^{(b)}$ and head set $\mathcal{D}^{(h)}$. In the $bb$ and $hh$ cases, we want to mitigate the influence of negative samples that may be geometrically distant. To achieve this, we introduce a distance-aware weighting penalty $e^{d_{i j}}$, where $d_{i,j}$ signifies the distance between respective bounding boxes $i$ and $j$. By combining these components, we can define the pulling loss as follow:
\begin{equation}
    L_{pull}=\mu (L_{pull}^{bb}+L_{pull}^{hh}) + \beta L_{pull}^{bh}
\end{equation}
In practical implementation, we set $\mu$ to 1.0 and $\beta$ to 1.5.
\par \noindent \textbf{Pushing Loss:} To ensure that the distance between the embedding vectors of negative pairs are as large as possible, we also design pushing losses for various cases: body and body ($bb$), head and head ($hh$), and matched body and head ($bh$). However, if the distance between the feature vectors exceeds a threshold $\sigma$, we consider the pair as an "easy" negative pair and exclude it from further processing. The pushing loss functions for three cases are as follow:
\begin{equation}
\left\{
\begin{array}{l}
L_{push}^{bb}=\frac{1}{M^{2}}\sum\limits_{i=1}^{M}\sum\limits_{j=1, j \neq i}^{M} \| \operatorname{max}(0, \delta -\mathbf{e}_{i}^{(b)}\mathbf{e}_{j}^{(b)}) \|^{2} \\
L_{push}^{hh}=\frac{1}{N^{2}}\sum\limits_{i=1}^{N}\sum\limits_{j=1, j \neq i}^{N} \| \operatorname{max}(0, \delta -\mathbf{e}_{i}^{(h)} \mathbf{e}_{j}^{(h)}) \|^{2}  \\
L_{push}^{bh}=\frac{1}{M} \frac{1}{N}\sum\limits_{i=1}^{M}\sum\limits_{j=1}^{N} \| \operatorname{max}(0, \delta -\mathbf{e}_{i}^{(b)} \mathbf{e}_{j}^{(h)}) \|^{2}  
\end{array}
\right.
\end{equation}
where $\sigma$ is the threshold (which we set to 2 by default), and $M$ and $N$ are similar to the settings in the pulling loss. By combining these components, we can define the pushing loss as follow:
\begin{equation}
    L_{push}=\mu (L_{push}^{bb}+L_{push}^{hh}) + \beta L_{push}^{bh}
\end{equation}
The the weights $\mu$ and $\beta$ in the pushing loss are the same as those used in the pulling loss.
\par \noindent \textbf{Associative Embedding Loss:} Given the pulling loss $L_{pull}$ and pushing loss $L_{push}$, we can obtain the Associative Embedding Loss by combining them with weighting coefficients $\sigma$ and $\tau$ as follow:
\begin{equation}
    Loss_{AML} = \sigma L_{pull} + \tau L_{push}
\end{equation}

\subsection{Dynamic Association}
\label{sec:dynamic_association}
\par With the approach outlined in dynamic detection part, we are able to obtain matched head and body detections $\mathcal{D}$. It is worth to note that certain bodies or faces may be absent, which can be represented as $\varnothing$.
\begin{equation}
   \mathcal{D}=\{
   (d_{0}^{(b)}, d_{0}^{(h)}),
   (d_{1}^{(b)}, \varnothing),
   (\varnothing, d_{2}^{(h)}),
   \dots,
   (d_{n}^{(b)}, d_{n}^{(h)})\}
\end{equation}

\par To effectively utilize the information from both head and body detections, we propose a novel dynamic association algorithm based on the idea of cascade matching. 
Cascade matching is a technique employed in DeepSORT~\cite{wojke2017deepsort} to facilitate the matching of historical frames.
In our dynamic association algorithm, we devise three distinct cases: matched head and body, mismatched body, and mismatched head. By employing the dynamic association algorithm for each of these cases, we are able to effectively leverage the information from both the head and body in occluded environments. Firstly, we associate the matched body and head detection boxes with tracklets to preserve comprehensive information. Subsequently, we merge mismatched body detection boxes with unmatched tracklets to establish a solid foundation for pedestrian tracking. Finally, we reconcile unmatched head detection boxes with unmatched tracklets, enabling us to effectively recover highly obscured objects.
The pseudo-code of Dynamic association is shown in Algorithm \ref{code:association}.
\begin{algorithm}[t]
	\caption{Dynamic association for head-body detections}
	\label{code:association}
	\begin{algorithmic}[1] 
		\Require Body and head detections $\mathcal{D}$
		\Ensure Tracks $\mathcal{T}$ of the video

            \Function {Dynamic Association}{$\mathcal{D}$}
                \State Initialization: $\mathcal{T} \leftarrow \varnothing$
                \For{frame $f_k$ in video $\mathcal{V}$}
                    \State $\mathcal{D}^{(bh)}, \mathcal{D}^{(b)}, \mathcal{D}^{(h)} \leftarrow \mathcal{D}$ 
                    \Statex   \LeftComment{2} {\texttt{/* matched body and head */}}
                    \State $\mathcal{T}_{remain}, {D}_{remain}^{(bh)} \leftarrow $ Asso($\mathcal{T}$, $\mathcal{D}^{(bh)}$)
                    \Statex   \LeftComment{2} {\texttt{/* mismatched body */}}
                    \State $\mathcal{T}_{remain}, {D}_{remain}^{(b)} \leftarrow $ Asso($\mathcal{T}_{remain}$, $\mathcal{D}^{(b)}$) 
                    \Statex   \LeftComment{2} {\texttt{/* mismatched head */}}
                    \State $\mathcal{T}_{remain}, {D}_{remain}^{(h)} \leftarrow $ Asso($\mathcal{T}_{remain}$, $\mathcal{D}^{(h)}$)
		          \Statex   \LeftComment{2} {\texttt{/* delete unmatched tracks */}}
		          \State $\mathcal{T} \leftarrow \mathcal{T} - \mathcal{T}_{remain}$
		          \Statex   \LeftComment{2} {\texttt{/* initialize new tracks */}}
		          \State $\mathcal{T} \leftarrow \mathcal{T} + \mathcal{D}_{remain}^{(bh)} + \mathcal{D}_{remain}^{(b)}$
                \EndFor
            \EndFunction

            \Function {Asso}{$\mathcal{T}$, $\mathcal{D}$}
                \State $\mathcal{D}_{high}, \mathcal{D}_{low} \leftarrow \mathcal{D}$ 
                \State Associate $\mathcal{T}$ and $\mathcal{D}_{high}$
    		\State $\mathcal{D}_{remain} \leftarrow$ remaining object boxes from $\mathcal{D}_{high}$
    		\State $\mathcal{T}_{remain} \leftarrow$ remaining tracks from $\mathcal{T}$
                \State \Return{$\mathcal{T}_{reamin}$, $\mathcal{D}_{remain}$}
            \EndFunction
	\end{algorithmic}
\end{algorithm}

\par The input of the dynamic association algorithm consists of a video sequence, denoted as $\mathcal{V}$, along with matched body and head detection boxes, represented by $\mathcal{D}$. It is important to note that the presence of occlusion may result in the absence of either the head or body in the detection. 
The objective of this algorithm is to output tracks, denoted as $\mathcal{T}$, for each object in the video. Each track contains the bounding box coordinates and identity of the object in each frame.
To achieve this, we first divide all the detection boxes in each frame into three categories: $\mathcal{D}^{(bh)}$ represents the matched body and head, $\mathcal{D}^{(b)}$ represents the mismatched body, and $\mathcal{D}^{(h)}$ represents the mismatched head. Once the detection boxes are separated, we apply the association function to each category to associate the detection boxes with their corresponding tracks.

\par The initial association is carried out between the matched body-head detection boxes and all the tracks in $\mathcal{T}$. The similarity matrix are computed using the Intersection over Union (IOU) and Re-ID feature distances between the matched detection boxes $\mathcal{D}^{(bh)}$ and the predicted boxes of tracks $\mathcal{T}$. The Hungarian Algorithm is then employed to complete the matching process based on the similarity matrix. Any unmatched detections are stored in ${D}_{remain}^{(bh)}$ and the unmatched tracks are stored in $\mathcal{T}_{remain}$.
Next, the second association is performed between the mismatched body detection boxes ${D}_{remain}^{(b)}$ and the remaining tracks $\mathcal{T}_{remain}$ after the first association. Similarity metrics are computed in the same manner as the first association, and the Hungarian Algorithm is applied for the second matching. The unmatched detections are saved in ${D}_{remain}^{(b)}$, and the unmatched tracks from the second association are stored in $\mathcal{T}_{remain}$.
Finally, the third association takes place between the mismatched head detection boxes ${D}_{remain}^{(h)}$ and the remaining tracks $\mathcal{T}_{remain}$ after the second association. The matching process is conducted in the same way as described above. Any unmatched detections are kept in ${D}_{remain}^{(h)}$ and the unmatched tracks from the third association are stored in $\mathcal{T}_{remain}$.

\par After the association, the unmatched tracks will be deleted from the tracklets. For each track in the unmatched tracks $\mathcal{T}_{remain}$ after the third association, only when it exists for more than a certain number of frames, i.e. 10, we delete it from the tracks $\mathcal{T}$. 
Finally, we initialize new tracks from the unmatched detection boxes ${D}_{remain}^{(bh)}$ and ${D}_{remain}^{(b)}$ after the third association. It is worth noting that we did not consider unmatched head detections ${D}_{remain}^{(h)}$, since heads are only used as supplementary information for tracking purposes, and introducing them into the main tracking framework would introduce more noise. 
As a result, the output of each individual frame will consist of the bounding boxes and identities of the tracks $\mathcal{T}$ in the current frame.

\begin{figure*}[t]
    \includegraphics[width=0.98\textwidth]{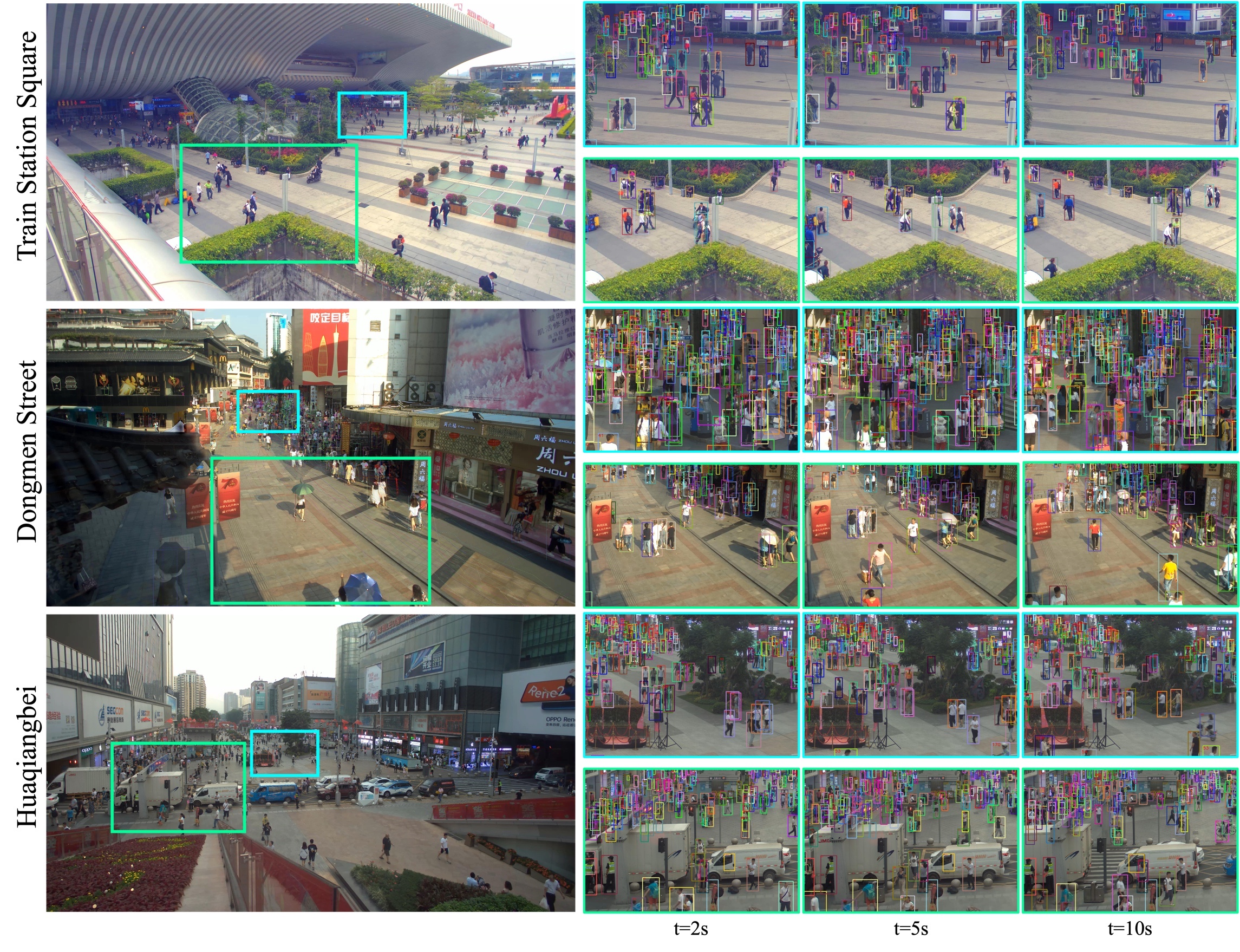}
    \centering
    \vspace{-3mm}
    \caption{Visualization results of DynamicTrack. We have selected gigapixel sequences from the test set of PANDA to demonstrate the effectiveness of DynamicTrack in handling complex crowded scenarios. In our visualizations, we utilize customizable visualization windows represented by green and blue rectangles. Additionally, we use colors to indicate different identities, with the same bounding box color indicating the same identity. }
    \vspace{-1mm}
    \label{fig:results}
\end{figure*}
\begin{figure}[t]
    \includegraphics[width=0.48\textwidth]{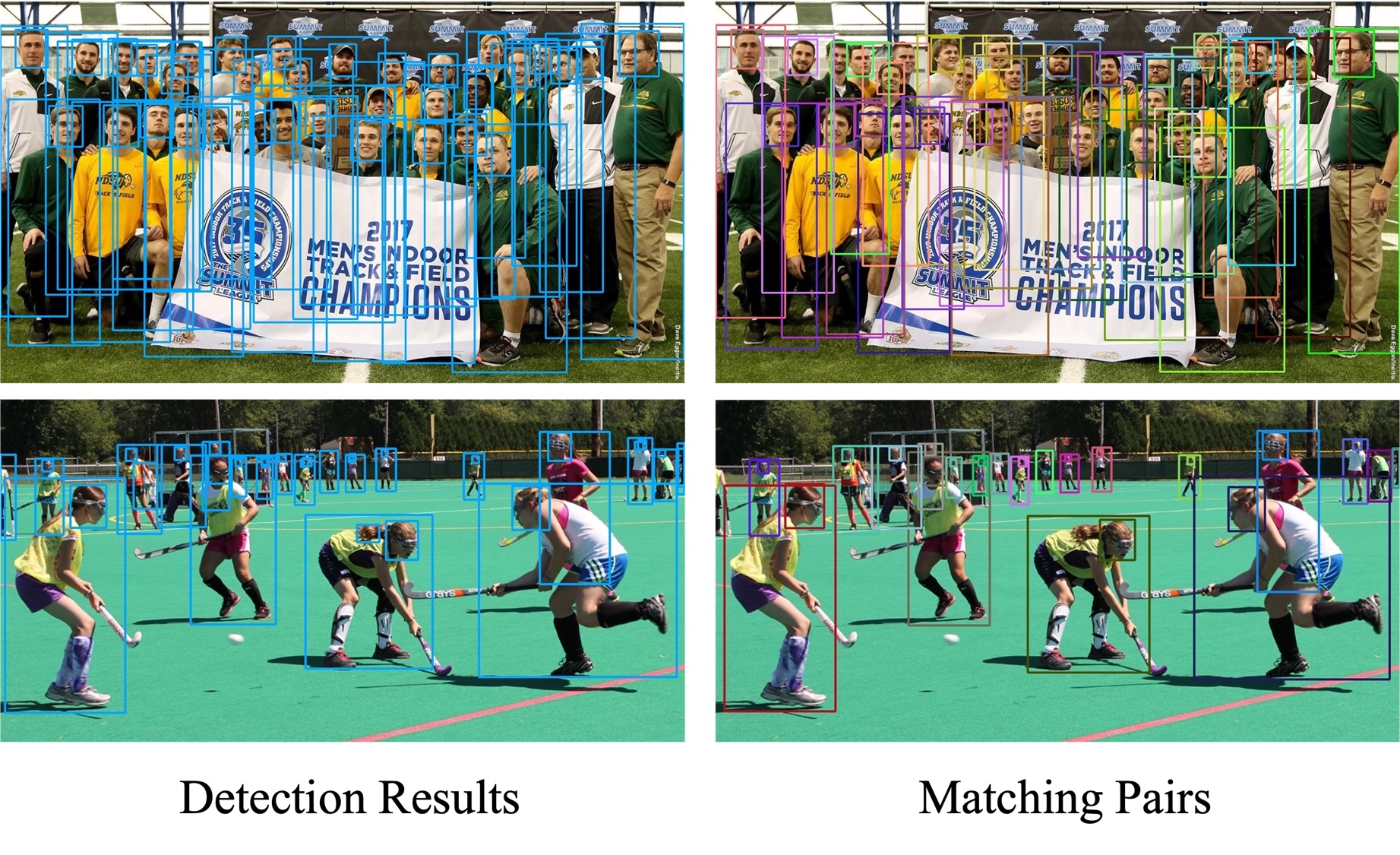}
    \centering
    \vspace{-3mm}
    \caption{Visualization of detection results and matching head and body pairs on CrowdHuman test set.}
    \label{fig:detector-vis}
    \vspace{-5mm}
\end{figure}

\begin{table}[t]
	\centering
 	\caption{Results on MOT20 val set. All trackers utilize the same detection results obtained from the dynamic detector.}
	\begin{tabular}{ccccc}
		\toprule
		Tracker  & MOTA$\uparrow$ & IDF1$\uparrow$ & HOTA$\uparrow$ &  IDs$\downarrow$ \\
		\midrule
		ByteTrack     & 68.5 & 71.4 & 57.6 & 3942 \\
		OCSORT        & 68.3 & 68.9 & 56.1 & 4037 \\
		StrongSORT    & 67.7 & 69.7 & 56.8 & 3253 \\
		BotSORT       & 69.4 & 71.8 & 57.7 & 3168 \\
		DynamicTrack  & 70.2 & 72.1 & 57.9 & 3376  \\
		\bottomrule
	\end{tabular}
	\label{tab:mot20-baseline}
        \vspace{-0.5mm}
\end{table}
\begin{table}[t]
	\centering
	\caption{Results on PANDA test set. All trackers utilize the same detection results obtained from the dynamic detector.}
	\begin{tabular}{ccccc}
		\toprule
		Tracker  & MOTA$\uparrow$ & IDF1$\uparrow$ & HOTA$\uparrow$ &  IDs$\downarrow$ \\
		\midrule
		ByteTrack     & 39.3 & 34.6 & 33.6 & 26265 \\
		OCSORT        & 43.3 & 34.8 & 36.0 & 44634 \\
		StrongSORT    & 55.8 & 53.9 & 48.1 & 11478 \\
		BotSORT       & 57.3 & 57.4 & 50.8 & 10235 \\
		DynamicTrack  & 60.4 & 59.2 & 53.9 & 7646  \\
		\bottomrule
	\end{tabular}
	\label{tab:panda-baseline}
        \vspace{-0.5cm}
\end{table}

\section{Experiment and Result}

\subsection{Experimental Setup}
We construct dynamic detector based on the well-known Faster-RCNN\cite{ren2015faster} architecture. We train the dynamic detector using the CrowdHuman dataset \cite{shao2018crowdhuman} and the same training weights were utilized for subsequent experiments. CrowdHuman is primarily focused on pedestrian detection in crowded scenes and provides precise annotations for both human body and head. To evaluate the association module, we conduct joint head and body tracking experiments on the widely-used MOT20 dataset, which includes challenging crowded scenes. Furthermore, we assess the performance of our DynamicTrack on the PANDA dataset \cite{wang2020panda}. The PANDA dataset is a gigapixel multi-object tracking dataset specifically designed for highly crowded and challenging scenes. 
When evaluating the detection performance, we utilize two widely used metrics: AP and $\text{MR}^{-2}$. For assessing the body-face association performance, we employ $\text{mMR}^{-2}$, a metric proposed in \cite{wan2021bfjdet}. This metric quantifies the proportion of body-face pairs that are miss-matched. 
When evaluating the tracking performance, we primarily rely on three widely-uesd evaluation metrics: MOTA, IDF1, and HOTA. MOTA predominantly assesses detection performance, while IDF1 emphasizes association performance. HOTA aims to strike a balance between accurate detection and association effects.

\subsection{Results of DynamicTrack}
\par \noindent \textbf{Tracking performance on MOT20.} In ~\cref{tab:mot20-baseline}, we provide the tracking results on the MOT20 dataset and compare them with the state-of-the-art two-stage methods on MOT20. We use the same detector trained on the CrowdHuman dataset as a baseline.  Obviously, our method achieves higher performance of MOTA which is the primary evaluation metric. 
\par \noindent \textbf{Tracking performance on PANDA.} 
In ~\cref{tab:panda-baseline}, we provide the gigapixel tracking results on the PANDA dataset. Our approach, DynamicTrack, is compared with state-of-the-art two-stage trackers including motion-based methods like ByteTrack\cite{zhang2022bytetrack} and OCSORT\cite{cao2023ocsort}, as well as appearance-based methods like BotSORT\cite{aharon2022botsort} and StrongSORT\cite{du2023strongsort}.
From the results, it can be observed that DynamicTrack achieves comparable performance to other state-of-the-art methods. 
\par \noindent \textbf{Qualitative results.} ~\cref{fig:results} showcases the tracking results under gigapixel sequences. Analysis from the results obtained on Trian Station Square, Dongmen Street and Huaqiangbei indicate that accurate tracking can be achieved in crowded scene of gigapixel sequences. This includes successful tracking of both sparse, large targets in the foreground as well as dense, small targets in the background.

\subsection{Ablation Study}
\par \noindent \textbf{Ablation study of Dynamic Detection.}
~\cref{tab:detector-results} presents the detection results on the CrowdHuman dataset. In this table, we compare our novel embedding-based method with the traditional position-based method. The position-based method calculates the IOU distance between the head and body detections, and then utilizes the Hungarian algorithm to select the best matching pairs. From the results, it is evident that our dynamic detector outperforms the position-based approach by a significant margin of \textbf{14.22\%} with respect to $\text{mMR}^{-2}$. Moreover, our dynamic detector maintains competitive detection performance. ~\cref{fig:detector-vis} showcases the detection results and matching results in crowded scenarios. It illustrates that our embedding-based method performs better, especially in situations involving complex occlusions. 

\par \noindent \textbf{Ablation study of Dynamic Association.}
To evaluate the impact of incorporating head information for tracking, we conducted experiments on the widely used MOT20 and PANDA dataset, which consist of crowded scenes. Specifically, we compare the effects of body-based tracking and body-head tracking. The results are summarized in ~\cref{tab:detector-ablation} and ~\cref{tab:panda-ablation}. From the results, it is clear that the inclusion of head information in tracking yields notable improvements. In MOT20 dataset, the head-body tracking method outperforms the body-based tracking method by 1.9 in terms of MOTA. Similarly, in PANDA dataset, the head-body tracking method surpasses the body-based tracking method by 4.7 in terms of MOTA. These results indicate that introducing head information can lead to significant performance gains, particularly in occluded environments.

\section{Conclusion}
In this paper, we address the challenging task of gigapixel tracking in crowded scenes by introducing DynamicTrack. To enhance the robustness to occlusion, we incorporate head information in addition to the traditional body-based features and leverage contrastive learning for dynamic detection. Moreover, we propose dynamic association algorithms for body-head tracking to overcome the challenges posed by gigapixel crowded sequences. Experimental results on benchmark MOT20 and PANDA have shown that our approach outperforms the state-of-the-art trackers in crowded scenes. The future plan is to integrate the dynamic detection framework into the latest transformer-based detectors to further enhance the tracking performance in crowded scenes.

\begin{table}[t]
	\centering
	\caption{Ablation study of the different modules of DynamicTrack on MOT20 test set.}
	\begin{tabular}{cc|cccc}
		\toprule
		Body & Head & MOTA$\uparrow$ & IDF1$\uparrow$ & HOTA$\uparrow$ & IDs\\
		\midrule
		\checkmark & & 68.3 & 71.3 & 57.4  & 4059\\
		\checkmark & \checkmark & 70.2 & 72.1 & 57.9 & 3376\\
		\bottomrule
	\end{tabular}
	\label{tab:detector-ablation}
\end{table}
\begin{table}[t]
	\centering
	\caption{Ablation study of the different modules of DynamicTrack on PANDA test set.}
	\begin{tabular}{cc|cccc}
		\toprule
		Body & Head & MOTA$\uparrow$ & IDF1$\uparrow$ & HOTA$\uparrow$ & IDs$\downarrow$\\
		\midrule
		  \checkmark &  & 55.7 & 55.3 & 48.5 & 10384\\
		\checkmark & \checkmark & 60.4 & 59.2 & 53.9 & 7646 \\
		\bottomrule
	\end{tabular}

	\label{tab:panda-ablation}
\end{table}

\begin{table}[t]
	\centering
	\caption{Results on the CrowdHuman validation set. Pos: Position-based method. Emb: Embedding-based method.}
	\begin{tabular}{c|ccc|c}
		\toprule
		Method & Class & $\text{AP}\uparrow$ & $\text{MR}^{-2}\downarrow$ & $\text{mMR}^{-2}\downarrow$ \\
		\midrule
		  \multirow{2}*{Emb}  & Head & 0.727 & 0.557 & \multirow{2}*{56.57} \\
		  ~                    & Body & 0.867 & 0.459 & ~ \\
            \midrule
		\multirow{2}*{Pos}  & Head & 0.743 & 0.532 & \multirow{2}*{70.79} \\
		~                    & Body & 0.867 & 0.441 & ~ \\
		\bottomrule
	\end{tabular}
	\label{tab:detector-results}
\end{table}

\bibliographystyle{IEEEtran}
\bibliography{sections/Reference}

\end{document}